\documentclass[sigconf]{acmart}

\usepackage{multirow}
\usepackage{booktabs}
\usepackage{amsmath,amsfonts}
\usepackage{algorithmic}
\usepackage{algorithm}
\usepackage{array}
\usepackage{url}
\usepackage{graphicx}
\usepackage{lipsum} 
\usepackage{colortbl}
\usepackage{booktabs}
\usepackage{hhline}
\AtBeginDocument{%
  }

\copyrightyear{2025}
\acmYear{2025}
\setcopyright{acmlicensed}
\acmConference[MM '25]{Proceedings of the 33rd ACM International Conference on Multimedia}{October 27--31, 2025}{Dublin, Ireland}
\acmBooktitle{Proceedings of the 33rd ACM International Conference on Multimedia (MM '25), October 27--31, 2025, Dublin, Ireland}
\acmDOI{XXXXXXX.XXXXXXX}
\acmISBN{978-1-4503-XXXX-X/2025/10}
\begin{document}

\title{Dual Prompt Learning for Adapting Vision-Language Models to Downstream Image-Text Retrieval}


\author{Yifan Wang}
\orcid{0009-0004-5607-3265}
\email{yifanwang1220@gmail.com}
\affiliation{
 \institution{College of Computer Science,\\ Sichuan University} 
 \city{Chengdu} 
 \country{China}
 }
\additionalaffiliation{\institution{Engineering Research Center of Machine Learning and Industry Intelligence, Ministry of Education, Chengdu, China}}

\author{Tao Wang}
\orcid{0000-0002-2480-878X}
\email{twangnh@gmail.com}
\authornotemark[1]
\authornote{Corresponding author.}
\affiliation{
 \institution{College of Computer Science,\\ Sichuan University} 
 \city{Chengdu} 
 \country{China}
 }

\author{Chenwei Tang}
\orcid{0000-0002-1749-986X}
\email{tangchenwei@scu.edu.cn}
\authornotemark[1]
\affiliation{
 \institution{College of Computer Science,\\ Sichuan University} 
 \city{Chengdu} 
 \country{China}
 }

\author{Caiyang Yu}
\orcid{0000-0001-8246-1561}
\email{yucy324@gmail.com}
\authornotemark[1]
\affiliation{
 \institution{College of Computer Science,\\ Sichuan University} 
 \city{Chengdu} 
 \country{China}
 }

\author{Zhengqing Zang}
\orcid{0009-0008-7904-9312}
\email{2022223045158@stu.scu.edu.cn}
\authornotemark[1]
\affiliation{
 \institution{College of Computer Science,\\ Sichuan University} 
 \city{Chengdu} 
 \country{China}
 }

\author{Mengmi Zhang}
\orcid{0000-0002-2694-7097}
\email{mengmi.zhang@ntu.edu.sg}
\affiliation{
  \institution{College of Computing and Data Science, \\Nanyang Technological University}
  \country{Singapore}
}
\additionalaffiliation{\institution{Deep NeuroCognition Lab, I2R and CFAR, Agency for Science, Technology and Research, Singapore}}

\author{Shudong Huang}
\orcid{0000-0001-6848-5460}
\email{huangsd@scu.edu.cn}
\authornotemark[1]
\affiliation{
 \institution{College of Computer Science,\\ Sichuan University} 
 \city{Chengdu} 
 \country{China}
 }

\author{Jiancheng Lv}
\authornotemark[1]
\orcid{0000-0001-6551-3884}
\email{lvjiancheng@scu.edu.cn}
\affiliation{
 \institution{College of Computer Science,\\ Sichuan University} 
 \city{Chengdu} 
 \country{China}
 }

\renewcommand{\shortauthors}{Yifan Wang et al.}

\begin{abstract}
\label{sec:ab}
Recently, prompt learning has achieved remarkable success in adapting pre-trained Vision-Language Models (VLMs) to downstream tasks such as image classification. However, its application to the downstream Image-Text Retrieval (ITR) task is more challenging. We find that the challenge lies in discriminating both fine-grained attributes and similar subcategories of the downstream data. To address this challenge, we propose \textbf{D}ual prompt Learning with Joint \textbf{C}ategory-\textbf{A}ttribute \textbf{R}eweighting (DCAR), a novel dual-prompt learning framework to achieve precise image-text matching. The framework dynamically adjusts prompt vectors from both semantic and visual dimensions to improve the performance of CLIP on the downstream ITR task. Based on the prompt paradigm, DCAR jointly optimizes attribute and category features to enhance fine-grained representation learning. Specifically, (1) at the attribute level, it dynamically updates the weights of attribute descriptions based on text-image mutual information correlation; and (2) at the category level, it introduces negative samples from multiple perspectives with category-matching weighting to learn subcategory distinctions. To validate our method, we construct the Fine-class Described Retrieval Dataset (FDRD), which serves as a challenging benchmark for ITR in downstream data domains. It covers over 1,500 downstream fine categories and 230,000 image-caption pairs with detailed attribute annotations. Extensive experiments on FDRD demonstrate that DCAR achieves state-of-the-art performance over existing baselines. 
The code and data are available at \url{https://github.com/wyf202322/DCAR}.
\end{abstract}




\maketitle

\section{Introduction}
\label{sec:intro}
\begin{figure}[t]
\begin{center}
    \includegraphics[width=1\linewidth]{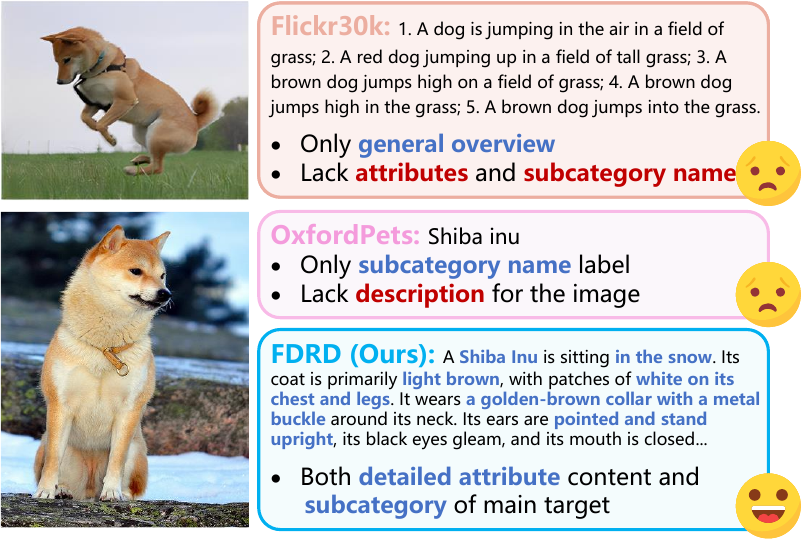}
\end{center}
\caption{Comparison between our dataset and existing datasets. The text inside the rounded rectangles represents the image captions. Our FDRD includes both detailed attribute descriptions and fine-grained category information.}
\label{fg: dataset_compare}
\end{figure}

With the development of Contrastive Language-Image Pre-training (CLIP)~\cite{CLIP}, pre-trained Visual Language Models (VLMs) have gained significant attention for their outstanding zero-shot recognition performance. 
By learning cross-modal representations from large-scale image-text pairs, CLIP builds a rich and unified visual-semantic space, yet its application to downstream data domains is sensitive to the trivialities in prompt engineering~\cite{CoOp,jin2021good}. 
Prompt learning has emerged as a powerful paradigm for adapting pre-trained VLMs to downstream tasks~\cite{CoCoOp,MaPLe,ge2023domain,ju2022prompting,lin2023being}, particularly in image classification. It introduces learnable prompts without modifying the original model parameters, preserving pre-trained knowledge while maximizing model potential using only a few labeled images. 

Despite its success in downstream classification tasks, the exploration of prompt learning remains limited in Image-Text Retrieval (ITR), a more fine-grained visual-semantic alignment task. Our preliminary analysis shows that the key challenge lies in learning prompts that facilitate both
\textbf{capturing fine-grained attributes} and \textbf{discriminating similar subcategories} in the downstream retrieval data.
For example, distinguishing the Bengal cat or Bombay cat while simultaneously recognizing detailed cat attributes such as color and pose~\cite{pan2023fine,yu2025rethinking}.
The challenge highlights the need for more specialized strategies when applying prompt learning to complex downstream retrieval scenarios, such as those in multimedia search~\cite{CLIP-Branches,zang2024zero} and recommendation systems~\cite{rajput2023recommender, naghiaei2022cpfair}.

To address the challenge, we propose \textbf{D}ual-prompt Learning with Joint \textbf{C}ategory-\textbf{A}ttribute \textbf{R}eweighting (DCAR), a prompt learning framework to achieve precise image-text retrieval.
Building on a prompt learning framework with both semantic and visual prompt vectors, DCAR jointly optimizes attribute and subcategory features to enhance fine-grained representation learning. (1) At the attribute-aware level, it employs mutual information to measure text-image correlation and dynamically adjusts description weights when calculating matching scores. This serves as an adaptive optimization scheme to enhance attribute identification. 
(2) At the category-aware level, it emphasizes fine-grained discrimination among subcategories of the same meta-category through negative sample augmentation with category-matching weighting. 
These help DCAR to boost its awareness of fine-grained features and adapt CLIP to a wide range of challenging retrieval scenarios.

\begin{figure}[t]
\begin{center}
    \includegraphics[width=1\linewidth]{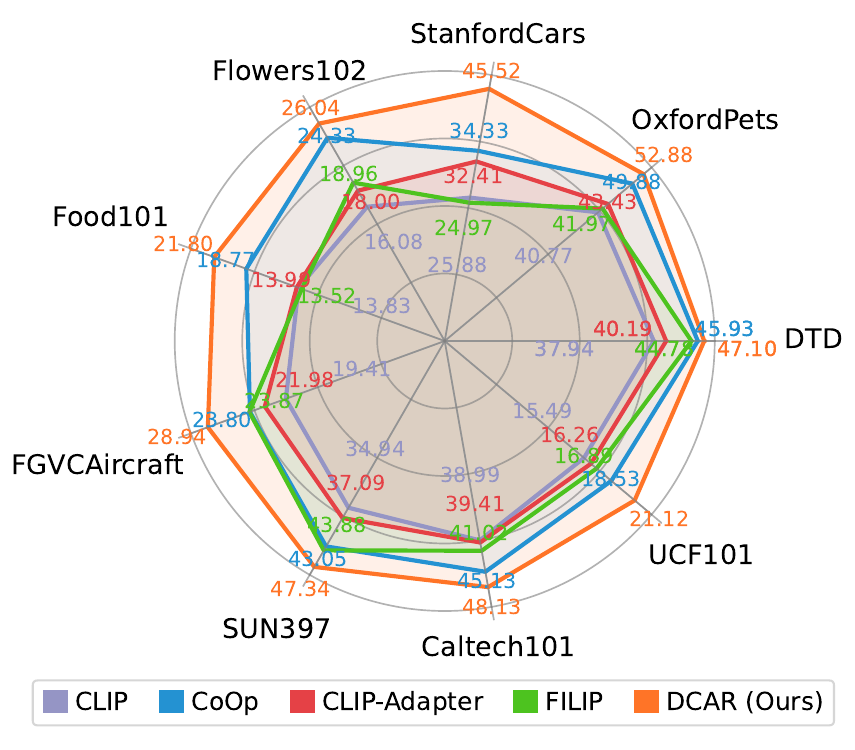}
\end{center}
\caption{Performance comparison. We evaluated the Image-to-Text Recall@1 results of different methods on FDRD. The results indicate that our FDRD dataset provides a diverse and challenging scenario for downstream retrieval tasks.}
\label{fg: yushiyan}
\end{figure}

In addition, current datasets for validating prompt learning mainly focus on downstream classification tasks, whereas suitable ITR benchmarks that can simultaneously capture fine-grained attributes and subtle category distinctions are still lacking. To bridge this gap, we construct the Fine-class Described Retrieval Dataset (FDRD), a more challenging ITR dataset featuring annotated captions that emphasize attributes and category distinctions. 
Compared to existing retrieval datasets~\cite{plummer2015flickr30k,pets}, our dataset covers a wide range of downstream domains, derived from traditional fine-grained datasets~\cite{fei2004learning,pets,cars,nilsback2008automated,bossard2014food,maji2013fine,xiao2010sun,cimpoi2014describing,soomro2012ucf101}, containing over \textbf{1,500} fine categories and \textbf{230,000} image-caption pairs (Fig.~\ref{fg: dataset_compare}). 
We will publish these fine caption annotations to facilitate future research. Following prior research~\cite{CoOp,CoCoOp}, we set up few-shot prompt learning with FDRD to evaluate image-text retrieval. Our experiments (Fig.~\ref{fg: yushiyan}) reveal that existing CLIP-based prompt learning methods~\cite{CLIP,CoOp} and retrieval approaches~\cite{Clip-adapter,FILIP} struggle with both fine-grained attribute recognition and subtle inter-subcategory discrimination.
Extensive experiments demonstrate that DCAR outperforms strong retrieval baselines and serves as a new baseline for studying the challenging downstream image-text retrieval task.

The key contributions of our work are as follows:
\begin{itemize}
    \item We identify the challenging task of adapting pre-trained vision-language models to downstream image-text retrieval, which entails aligning both fine-grained semantic object categories and visual attributes. 
\end{itemize}
\begin{itemize}
    \item We propose DCAR, a dual-prompt learning framework that enhances CLIP for downstream Image-Text Retrieval (ITR) via joint category-attribute adaptive learning. DCAR aims to improve retrieval precision with more effective distinctions on attributes and fine-grained categories. 
\end{itemize}
\begin{itemize}
    \item We construct the Fine-class Described Retrieval Dataset (FDRD) for the ITR task. It provides a more challenging benchmark featuring annotated captions that emphasize subtle attributes and subcategory distinctions.
\end{itemize}

\section{Related Work}
\label{sec: related}
\subsection{Fine-grained Image-Text Retrieval}
 Existing Fine-grained Image-Text Retrieval (FG-ITR) tasks can be broadly divided into two types: (1) classification-based retrieval~\cite{moskvyak2021keypoint,wang2023fine,wei2017selective,xie2023ra,park2024META}, which aims to retrieve images or texts at the fine-grained category level. It focuses on matching items in the same category based on a given query. However, it struggles to effectively utilize attribute-level information, significantly limiting its practical application in real-world scenarios; (2) caption-based retrieval~\cite{yang2023alip, vasu2024mobileclip,tschannen2023clippo,li2022image,jia2021scaling,lu2022cots}, which aims to match attribute information between image and text. It captures attributes like color, size, and shape for retrieval. But it is constrained by dataset limitations and lacks adaptability to diverse downstream tasks.
 Our DCAR enhances FG-ITR by jointly modeling fine-grained attributes and subcategory information, enabling more practical and precise image-text alignment.

\subsection{Prompt Learning}
Inspired by developments in Natural Language Processing (NLP)~\cite{LesterAC21,ZhongFC21,LiL20}, prompt learning has emerged as a powerful technique in VLMs~\cite{CLIP,yao2024cpt}. This approach enables the efficient adaptation of pre-trained models to downstream tasks without extensive retraining. The pioneering work CoOp~\cite{CoOp} introduced the concept of learning prompts by optimizing prompt vectors, demonstrating strong performance in few-shot settings for fine-grained classification tasks. 
VPT~\cite{VPT} adapts prompt learning to the visual domain by learning continuous vectors in the input space. MaPle~\cite{MaPLe} emerged, leveraging a dual-prompt architecture to better capture cross-modal interactions. Despite the success of prompt learning in classification tasks~\cite{park2024META, guo2023calip,Vt-clip,lee2023read,wang2023open}, its application to image-text retrieval remains underexplored. Our DCAR is the first work to introduce dual-prompt learning for achieving more precise retrieval. By constructing adaptive prompts for text and image modalities separately, our model can utilize pre-trained knowledge to improve retrieval performance more fully.

\begin{figure*}[t]
\begin{center}
    \includegraphics[width=1\linewidth]{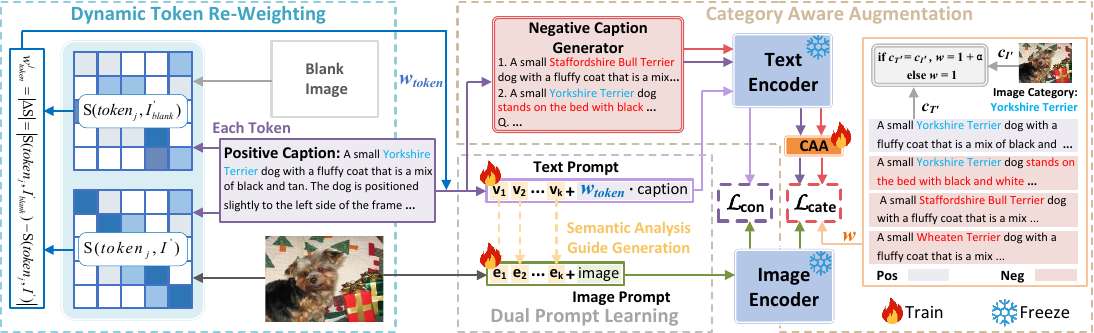}
\end{center}
\caption {Overview of DCAR approach. We introduce a dual-prompt learning framework to achieve precise image-text matching by integrating (1) an attribute-aware token re-weighting module and (2) a category-aware negative augmentation module.}
\label{fg: overview}
\end{figure*}

\section{Proposed Method}
\label{sec: method}
The overview of our method is shown in Fig.~\ref{fg: overview}. We present an adaptive prompt learning framework named \textbf{D}ual-prompt Learning with Joint \textbf{C}ategory-\textbf{A}ttribute \textbf{R}eweighting (DCAR).
Specifically, the framework introduces dual prompts to optimize both visual and textual representations and ensures robust cross-modal alignment with two key aspects: (1) \textbf{Dynamic Token Re-Weighting}: We dynamically re-weight tokens based on their mutual information with text-image correlations, enhancing attribute awareness. (2) \textbf{Category Aware Augmentation}: We employ negative sample augmentation with emphasis on the fine-grained discrimination among subcategories within the same meta-category. 
These components improve retrieval accuracy by aligning descriptions with visual semantics. 
We introduce these components sequentially in the following subsections.

\subsection{Preliminary}
CLIP (Contrastive Language-Image Pre-training~\cite{CLIP}) establishes a joint embedding space that aligns visual and textual representations through contrastive learning. 
Given an input image $\mathbf{x}$ and its corresponding text $\mathbf{t}$, CLIP processes them through the image encoder $f_{\text{img}}(\cdot)$ and the text encoder $f_{\text{text}}(\cdot)$, producing the image embedding $I = f_{\text{img}}(\mathbf{x})$ and the text embedding $T = f_{\text{text}}(\mathbf{t})$, respectively. The similarity score $\text{S}(\cdot)$ between image and text features is defined as:
\begin{equation}
\label{cos}
    \text{S}(I_i, T_i) = \exp\left(\frac{I_i \cdot T_i}{\left|I_i\right| \left|T_i\right|}\right),
\end{equation}
where $I_i$ and $T_i$ denote the $i$-th image and text embeddings, respectively. The contrastive loss is then computed as:
\begin{equation}
\label{eq:clip_loss}
\begin{aligned}
    \mathcal{L}_{\text{CLIP}} = -\frac{1}{N} \sum_{i=1}^N \left[
        \log \frac{\text{S}(I_i, T_i) / \tau}{\sum_{j=1}^N \text{S}(I_i, T_j) / \tau}+\log \frac{\text{S}(T_i, I_i) / \tau}{\sum_{j=1}^N \text{S}(T_i, I_j) \tau}\right],
\end{aligned}
\end{equation}
where $N$ is the batch size, and $\tau$ is a temperature parameter. The index $j$ denotes each possible candidates in the batch. 

With its cross-modal matching capability, CLIP can be leveraged to perform the image-text retrieval task. For example, given a query image $I_q$ and a set of texts $\{T_i\}_{i=1}^{N_e}$ in the database, the similarity score $\text{S}(I_q, T_i)$ between the query image and the $i$-th text can be calculated. The Top-$k$ candidates are then selected based on these similarity scores: 
\begin{equation}
R_{\text{Top-$k$}} = \text{Top-$k$} \left(\left\{\text{S}(I_q, T_i)\right\}_{i=1}^{N_e}\right),
\end{equation}
where $N_e$ denotes the total number of entries in the database, and the function $\text{Top-}k$ returns the $k$ most relevant results. The process of querying text $T_q$ is similar.

\subsection{Dual Prompt Learning}
\label{sub:Dual_Prompt}
 Applying the pre-trained CLIP model to the downstream image-text retrieval task is non-trivial due to the distribution shift in both image and text modalities. Recently, prompt learning has emerged as a highly effective paradigm for adapting Vision-Language Models (VLMs) to downstream tasks. Motivated by this, we develop a prompt learning approach by optimizing learnable visual and textual prompts while preserving the original model parameters. Specifically, prompt learning method such as CoOp~\cite{CoOp} typically constructs its prompt templates by combining class labels with trainable context vectors:
\begin{equation}
\label{eq: class prompts}
\mathcal{P}_{\text{class}} = \left(v_1, v_2, \dots, v_k, \langle\text{class}\rangle\right),
\end{equation}
where \(\mathcal{P}_{\text{class}}\) denotes the class-specific prompt, \(\{v_1, v_2, \dots, v_k\}\) denote the learnable context vectors, and $\langle\text{class}\rangle$ represents the tokens of the class name. The corresponding text feature is computed as:
\begin{equation}
T = f_{\text{text}}(\mathcal{P}_{\text{class}}).
\label{eq:text_embedding}
\end{equation}

\textbf{Text Prompt.} Inspired by CoOp, we construct the textual prompt template by combining text and learnable context vectors: 
\begin{equation}
\label{eq: caption prompts}
\mathcal{P}_{\text{text}} = \left(v_1, v_2, \dots, v_k, \langle\text{text}\rangle\right),\quad
T^{\prime}=f_{\text{text}}(\mathcal{P}_{\text{text}}),
\end{equation}
where \(\{v_1, v_2, \dots, v_k\}\) denote the learnable context vectors for the text, \(\mathcal{P}_{\text{text}}\) represents the text prompt, \(T^{\prime}\) denotes the text feature of caption, and $\langle\text{text}\rangle$ is the tokens of input text.

\textbf{Visual Prompt.} In VLMs, simply tuning text prompts is often insufficient to fully bridge the gap between pre-training and downstream classification tasks, leading to suboptimal performance~\cite{xing2023dual}. Motivated by prior work~\cite{MaPLe} that introduces a text-conditioned visual prompt for downstream classification tasks, we construct a visual prompt by extracting data-discriminative features through semantic analysis of the outputs from the text backbone network:
\begin{equation}
\label{eq:visual prompt }
\{e_1, e_2, \dots, e_k\}=  F(\{v_1, v_2, \dots, v_k\}),
\end{equation}
where $\{e_1, e_2, \dots, e_k\}\in \mathcal{R}^{d_e \times k}$ represents the learnable visual vectors. The function $F(\cdot)$ is a linear transformation layer that projects the learnable context vectors (dimension $d_v$) from the semantic space of text to the visual feature embedding space (dimension $d_e$).The visual prompt is defined as:
\begin{equation}
\label{eq:image_embedding }
\mathcal{P}_{\text{img}} = \left(e_1, e_2, \dots, e_k, \langle\text{image}\rangle\right),\quad
I^{\prime}=f_{\text{img}}(\mathcal{P}_{\text{img}}),
\end{equation}
where \(\mathcal{P}_{\text{img}}\) represents the visual prompt, \(I^{\prime}\) denotes the image feature, $\langle\text{image}\rangle$ represents the input image embedding.

With the weights of the image encoder $f_{\text{img}}(\cdot)$ and the text encoder $f_{\text{text}}(\cdot)$ frozen, only the visual and the textual vectors are optimized with the contrastive loss in Eq.~\ref{eq:clip_loss}.

\subsection{Dynamic Token Re-Weighting}
\label{sub:Re-Weighting}
Our preliminary experiments show that the complex captions in the Fine-class Described Retrieval Dataset (FDRD) pose a significant challenge for aligning fine-grained features. This requires learning fine-grained discriminative features and precise alignment of image-related textual elements, while minimizing the impact of abstract concepts or syntactic connectors. To address this, we use a dynamic token re-weighting method based on Mutual Information (MI) to improve attribute-level awareness. 

MI measures the dependence between two random variables~\cite{kraskov2004estimating,belghazi2018mine}. This measure is beneficial for cross-modality analysis, as the statistical features are assumed to be consistent. Previous studies have successfully applied MI in cross-modality tasks, including data retrieval~\cite{hoang2022multimodal,bachman2019learning}, representation learning~\cite{guo2022online,niu2024neural,wu2025enhancing}, and domain adaptation~\cite{cha2022domain,wu2024diffusion}. In our work, we employ MI, denoted as $\text{MI}( ; )$, to evaluate the correlation between a given token and the image:
\begin{equation}
\label{mi}
    \text{MI}(token_{j}; I^{\prime}) = H(token_{j}) - H(token_{j}|I^{\prime}),
\end{equation}
where $token_{j}$ denotes the $j$-th token embedding in the positive caption of the image, and $I^{\prime}$ is the image embedding. $H(token_{j})$ represents the entropy of the $j$-th token, and $H(token_{j}|I^{\prime})$ denotes the conditional entropy given the image. Tokens with high MI value indicate greater relevance to the target image and are assigned higher weight, whereas tokens with lower MI value receive reduced weighting due to their weaker correlation. We can reweigh each token based on its MI score, as:
\begin{equation}
\label{mi}
    w^{j}_{token} = |\text{MI}(token_{j}; I^{\prime})|.
\end{equation}

The MI can be viewed as a measure of the information shared between the token and the image. To approximate MI, we introduce a comparison between a target image, which contains meaningful content, and a blank image, which serves as an uninformative baseline. For a given token, we compute its embedding similarity with both the blank image and the target image (illustrated in the cyan dashed box of Fig.~\ref{fg: overview}) and define the similarity difference as:
\begin{equation}
    \Delta\text{S} = \text{S}(token_{j}, I^{\prime}_{\text{blank}}) - \text{S}(token_{j}, I^{\prime}),
\end{equation}
where \(I^{\prime}\) and \(I^{\prime}_{\text{blank}}\) represent the target and blank image embeddings, respectively. 

Based on Energy Based Models~\cite{liu2020energy,lecun2006tutorial}, it can be shown that MI between $token_{j}$ and the target image embedding $I^{\prime}$ is proportional to $\Delta\text{S}$ (refer to supplementary material for detailed proof):
\begin{equation}
    \text{MI}(token_{j}; I^{\prime}) \propto \Delta\text{S}.
\end{equation}

Thus, we approximate MI using the similarity difference between token embeddings and image embeddings. The larger the difference, the stronger the correlation between the token and the target image; the smaller the difference, the token is considered less relevant and is regarded as noise. Thus, we can rephrase the token weight in terms of similarity differences $\Delta\text{S}$:
\begin{equation}
\label{eq: ΔS}
    w^{j}_{token} = |\Delta\text{S}|.
\end{equation}

When calculating the text features, we assign greater weights to tokens that are highly correlated with the image, while tokens with lower correlation are given lower weights:
\begin{equation}
\label{eq: weight_T}
T^{\prime}_{w} = f_{\text{text}}\left( \left[ \tilde{w}^{1} \cdot token_1, \tilde{w}^{2} \cdot token_2, \dots, \tilde{w}^{n} \cdot token_n \right] \right),
\end{equation}
where $token_j$ denotes the $j$-th token in the positive caption. $\tilde{w}^{j}_{token} = {w^{j}_{token}}/{\sum_{k=1}^{n} w^{k}_{token}}$ denotes normalized weight for the token. By incorporating the similarity difference weights, we obtain an improved contrastive loss for retrieval:
\begin{equation}
\label{eq:con_loss2}
\begin{aligned}
\mathcal{L}_{\mathrm{con}}=-\frac{1}{N} \sum_{i=1}^N\left[
\log \frac{\text{S}\left(I^{\prime}_i, T^{\prime}_{wi}\right) / \tau}{\sum_{j=1}^{N}\text{S}\left(I^{\prime}_i, T^{\prime}_{wj}\right) / \tau}+ \log \frac{\text{S}\left(T^{\prime}_{wi}, I^{\prime}_i\right) / \tau}{\sum_{j=1}^{N}\text{S}\left(T^{\prime}_{wi}, I^{\prime}_j\right) / \tau}\right],
\end{aligned}
\end{equation}
where $I^{\prime}_i$ denotes the $i$-th image feature, and $T^{\prime}_{wi}$ is the $i$-th text feature after weighting. 

\subsection{Category Aware Augmentation}
\label{sub:Category}
To further emphasize the distinction between the fine categories, we incorporate multiple caption augmentations by generating $Q$ negative samples for each positive caption, which enhances the compositional reasoning in VLMs~\cite{neg_text_aug1,neg_text_aug2,yu2024gpt}. Negative captions are created from multiple perspectives by altering subcategory name words or modifying descriptions within the original caption. 

Building on this negative caption augmentation strategy, we introduce a category aware augmentation (CAA) function $F_{\text{CAA}}(\cdot)$ after the text encoder, which consists of two linear layers. A category-sensitive loss is then defined as:
\begin{equation}
\label{eq:cate_loss}
\mathcal{L}_{\text{cate}}=-\sum_{i=1}^N \log \frac{\text{S}\left(F_{\text{CAA}}(T^{\prime}_i), I^{\prime}_i\right)/ \tau}{\text{S}\left(F_{\text{CAA}}(T^{\prime}_i), I^{\prime}_i\right)/\tau+ \sum_{j=1}^Q\text{S}\left(F_{\text{CAA}}(T_j^{\text {neg}}), I^{\prime}_i\right)/ \tau},
\end{equation}
where \(T_j^{\text{neg}}\) represents the \(j\)-th negative caption sample feature, \(Q\) is the number of negative samples. \(I^{\prime}\) denotes image feature,\(T^{\prime}\) denotes the positive caption samples feature.

To avoid over-reliance on category names and ensure attention to other caption details, we introduce a category-sensitive weight to balance positive and negative samples based on their category matching (see Fig.~\ref{fg: overview}, top right).
This weight is adjusted according to the category matching of the samples, and it is defined as:
\begin{equation}
    w = 1 + \alpha \cdot \mathbb{I}(c_{T^{\prime}},c_{I^{\prime}}),
\end{equation}
where \(\alpha\) is a scaling factor, and \(\mathbb{I}(\cdot)\) is an indicator function, which takes a value of 1 when the category in input text $c_{T^{\prime}}$ matches the category in image $c_{I^{\prime}}$, and zero otherwise. 

When \(c_{T^{\prime}} = c_{I^{\prime}}\), indicating category matching, $\mathbb{I}(c_{T^{\prime}},c_{I^{\prime}})=1$, and the weight \(w = 1 + \alpha\), reinforcing the model to prioritize learning finer distinctions within matching categories. Conversely, when \(c_{T^{\prime}} \neq c_{I^{\prime}}\), $\mathbb{I}(c_{T^{\prime}},c_{I^{\prime}})=0$, the weight reduces to \(w = 1\), which avoids over-penalizing negatives and prevents the model from relying solely on category names for discrimination. 
Specifically, for positive captions, the weight is \(w = 1 + \alpha\). For negative samples, the weight is divided into two cases: \(w = 1\) for negative samples with a correct attribute description but an incorrect category, and \(w = 1 + \alpha\) for negative samples with a correct category but incorrect attribute description.

By incorporating the category-sensitive weight \(w\) into the loss function, we obtain an improved category-sensitive loss:
\begin{equation} \label{eq:cate_sent}
\begin{aligned}
&\mathcal{L}_{\text{cate}}= \\
&-\sum_{i=1}^N \log \frac{w^i \cdot \text{S}\left(F_{\text{CAA}}(T^{\prime}_i), I^{\prime}_i\right)/\tau}{w^i \cdot \text{S}\left(F_{\text{CAA}}(T^{\prime}_i), I^{\prime}_i\right)/\tau + \sum_{j=1}^Q w^{i,j} \cdot \text{S}\left(F_{\text{CAA}}(T^{\text{neg}}_{j}), I^{\prime}_i\right)/\tau},
\end{aligned}
\end{equation}
where $w^{i,j}$ denotes the weight of the \(j\)-th negative caption samples, and $w^{i}$ denotes the weight of the positive caption. 
The final total loss function is defined as:
\begin{equation}
\label{eq: total_loss}
\mathcal{L}_{\text {total}}=\lambda_1 \mathcal{L}_{\mathrm{con}}+\lambda_2 \mathcal{L}_{\text {cate}},
\end{equation}
where $\lambda_1$ and $\lambda_2$ are hyperparameters.

\begin{figure}[t]
\begin{center}
    \includegraphics[width=1\linewidth]{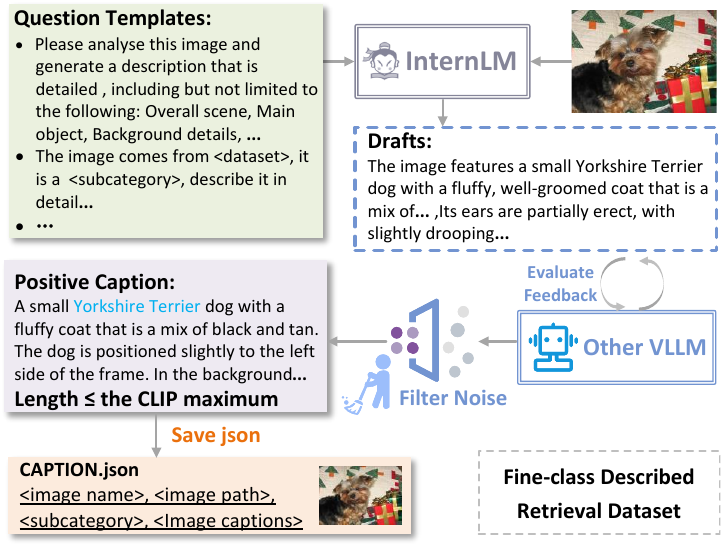}
\end{center}
\caption {The pipeline for constructing the Fine-class Described Retrieval Dataset (FDRD).}
\label{fg: construct_data}
\end{figure}

\begin{table*}
\centering
\caption{Performance comparison on the Fine-class Described Retrieval Dataset. The model is trained under a 16-shot setting and evaluated on the test sets. AVG: average performance across downstream fine-grained domains.}
\resizebox{\textwidth}{!}{
\begin{tabular}{lc|cccccccc|cccccccc} 
\toprule
\multicolumn{1}{c}{\multirow{2}{*}{Dataset}}                            & \multirow{2}{*}{R@K} & \multicolumn{8}{c|}{Image-to-Text}                                                                              & \multicolumn{8}{c}{Text-to-Image}                                                                                                        \\ 
\hhline{~~----------------}
\multicolumn{1}{c}{}                                                    &                      & CLIP  & CoOp  & Adapter & MaPle & FILIP & Alip  & FineCLIP & {\cellcolor[rgb]{0.898,0.898,0.898}}Ours           & CLIP  & CoOp  & Adapter & MaPle & FILIP          & Alip           & FineCLIP       & {\cellcolor[rgb]{0.898,0.898,0.898}}Ours            \\ 
\midrule
\multirow{3}{*}{\begin{tabular}[c]{@{}l@{}}Caltech\\101\end{tabular}}   & 1                    & 38.99 & 45.13 & 39.41   & 46.26 & 41.02 & 43.51 & 46.46    & {\cellcolor[rgb]{0.898,0.898,0.898}}\textbf{48.13} & 32.17 & 29.70 & 30.22   & 29.69 & 33.23          & 35.86          & \textbf{37.20} & {\cellcolor[rgb]{0.898,0.898,0.898}}36.78           \\
                                                                        & 5                    & 69.49 & 76.12 & 70.93   & 77.09 & 70.43 & 74.23 & 77.75    & {\cellcolor[rgb]{0.898,0.898,0.898}}\textbf{80.14} & 64.54 & 60.81 & 60.44   & 60.97 & 65.21          & 68.94\textbf{} & \textbf{71.43} & {\cellcolor[rgb]{0.898,0.898,0.898}}70.16           \\
                                                                        & 10                   & 82.68 & 87.03 & 83.34   & 87.31 & 84.17 & 83.97 & 85.24    & {\cellcolor[rgb]{0.898,0.898,0.898}}\textbf{89.00} & 78.66 & 75.86 & 75.42   & 76.55 & 78.84          & 80.15          & 82.24          & {\cellcolor[rgb]{0.898,0.898,0.898}}\textbf{82.31}  \\ 
\midrule
\multirow{3}{*}{\begin{tabular}[c]{@{}l@{}}Oxford\\Pets\end{tabular}}   & 1                    & 40.77 & 49.88 & 43.43   & 49.40 & 41.97 & 47.11 & 48.45    & {\cellcolor[rgb]{0.898,0.898,0.898}}\textbf{52.88} & 30.69 & 31.59 & 29.19   & 30.50 & 31.28          & 36.63          & 35.93          & {\cellcolor[rgb]{0.898,0.898,0.898}}\textbf{39.82}  \\
                                                                        & 5                    & 70.13 & 78.81 & 72.83   & 78.36 & 72.73 & 75.60 & 75.88    & {\cellcolor[rgb]{0.898,0.898,0.898}}\textbf{81.80} & 55.52 & 59.61 & 54.21   & 57.94 & 57.38          & 61.01          & 66.78          & {\cellcolor[rgb]{0.898,0.898,0.898}}\textbf{69.56}  \\
                                                                        & 10                   & 81.60 & 87.80 & 83.36   & 87.63 & 83.79 & 83.68 & 85.17    & {\cellcolor[rgb]{0.898,0.898,0.898}}\textbf{90.93} & 68.38 & 72.14 & 67.08   & 71.16 & 69.77          & 76.81          & 80.43          & {\cellcolor[rgb]{0.898,0.898,0.898}}\textbf{81.08}  \\ 
\midrule
\multirow{3}{*}{\begin{tabular}[c]{@{}l@{}}Stanford\\Cars\end{tabular}} & 1                    & 25.88 & 34.33 & 32.41   & 43.50 & 24.97 & 28.12 & 40.78    & {\cellcolor[rgb]{0.898,0.898,0.898}}\textbf{45.52} & 18.60 & 21.63 & 18.65   & 26.78 & 16.14          & 21.04          & 27.20          & {\cellcolor[rgb]{0.898,0.898,0.898}}\textbf{32.90}  \\
                                                                        & 5                    & 52.22 & 65.62 & 63.26   & 76.44 & 51.63 & 58.95 & 67.46    & {\cellcolor[rgb]{0.898,0.898,0.898}}\textbf{79.05} & 43.22 & 46.81 & 43.24   & 54.96 & 41.29          & 47.82          & 59.18          & {\cellcolor[rgb]{0.898,0.898,0.898}}\textbf{64.06}  \\
                                                                        & 10                   & 65.96 & 78.10 & 75.87   & 86.17 & 63.25 & 71.48 & 80.31    & {\cellcolor[rgb]{0.898,0.898,0.898}}\textbf{88.41} & 56.78 & 59.62 & 55.79   & 67.45 & 53.31          & 61.81          & 67.82          & {\cellcolor[rgb]{0.898,0.898,0.898}}\textbf{76.21}  \\ 
\midrule
\multirow{3}{*}{\begin{tabular}[c]{@{}l@{}}Flowers\\102\end{tabular}}   & 1                    & 16.08 & 24.33 & 18.00   & 24.38 & 18.96 & 20.34 & 20.75    & {\cellcolor[rgb]{0.898,0.898,0.898}}\textbf{26.04} & 12.87 & 12.11 & 10.32   & 12.38 & 13.75          & 13.56          & 16.16          & {\cellcolor[rgb]{0.898,0.898,0.898}}\textbf{18.56}  \\
                                                                        & 5                    & 41.21 & 57.63 & 47.68   & 60.30 & 51.31 & 51.44 & 54.18    & {\cellcolor[rgb]{0.898,0.898,0.898}}\textbf{62.47} & 39.59 & 36.19 & 33.94   & 37.92 & 40.55          & 40.20          & 42.97          & {\cellcolor[rgb]{0.898,0.898,0.898}}\textbf{43.65}  \\
                                                                        & 10                   & 57.00 & 75.16 & 64.98   & 76.51 & 66.10 & 68.81 & 72.85    & {\cellcolor[rgb]{0.898,0.898,0.898}}\textbf{78.83} & 58.14 & 52.11 & 49.17   & 55.66 & 58.62          & 59.51          & 60.59          & {\cellcolor[rgb]{0.898,0.898,0.898}}\textbf{61.30}  \\ 
\midrule
\multirow{3}{*}{Food101}                                                & 1                    & 13.83 & 18.77 & 13.99   & 19.50 & 13.52 & 15.24 & 19.06    & {\cellcolor[rgb]{0.898,0.898,0.898}}\textbf{21.80} & 6.98  & 8.50  & 6.20    & 8.00  & 7.82           & 8.50           & 10.28          & {\cellcolor[rgb]{0.898,0.898,0.898}}\textbf{13.03}  \\
                                                                        & 5                    & 29.59 & 39.53 & 31.06   & 40.44 & 30.41 & 36.66 & 41.20    & {\cellcolor[rgb]{0.898,0.898,0.898}}\textbf{44.20} & 17.99 & 20.72 & 16.63   & 20.18 & 17.70          & 19.45          & 22.45          & {\cellcolor[rgb]{0.898,0.898,0.898}}\textbf{26.98}  \\
                                                                        & 10                   & 38.71 & 50.39 & 41.35   & 51.65 & 41.33 & 45.60 & 50.29    & {\cellcolor[rgb]{0.898,0.898,0.898}}\textbf{53.08} & 25.78 & 29.18 & 23.83   & 28.59 & 26.25          & 29.31          & 34.07          & {\cellcolor[rgb]{0.898,0.898,0.898}}\textbf{37.33}  \\ 
\midrule
\multirow{3}{*}{\begin{tabular}[c]{@{}l@{}}FGVC\\Aircraft\end{tabular}} & 1                    & 19.41 & 23.80 & 21.98   & 26.12 & 23.87 & 20.36 & 22.33    & {\cellcolor[rgb]{0.898,0.898,0.898}}\textbf{28.94} & 18.93 & 12.84 & 12.48   & 14.12 & 19.13          & 18.72          & \textbf{20.96} & {\cellcolor[rgb]{0.898,0.898,0.898}}20.47           \\
                                                                        & 5                    & 41.91 & 51.66 & 46.94   & 54.03 & 47.13 & 43.29 & 53.64    & {\cellcolor[rgb]{0.898,0.898,0.898}}\textbf{56.43} & 42.12 & 30.36 & 30.54   & 31.56 & 43.54          & 41.52          & 43.31          & {\cellcolor[rgb]{0.898,0.898,0.898}}\textbf{44.04}  \\
                                                                        & 10                   & 52.45 & 64.07 & 57.59   & 66.75 & 58.58 & 55.46 & 64.43    & {\cellcolor[rgb]{0.898,0.898,0.898}}\textbf{68.32} & 52.50 & 40.56 & 40.02   & 42.33 & 53.65          & 51.72          & 55.31          & {\cellcolor[rgb]{0.898,0.898,0.898}}\textbf{55.60}  \\ 
\midrule
\multirow{3}{*}{SUN397}                                                 & 1                    & 34.94 & 43.05 & 37.09   & 45.60 & 43.88 & 42.14 & 42.53    & {\cellcolor[rgb]{0.898,0.898,0.898}}\textbf{47.34} & 23.48 & 19.03 & 19.97   & 18.62 & 30.22          & 29.97          & 29.28          & {\cellcolor[rgb]{0.898,0.898,0.898}}\textbf{34.30}  \\
                                                                        & 5                    & 62.06 & 72.65 & 65.05   & 74.36 & 69.82 & 69.64 & 69.07    & {\cellcolor[rgb]{0.898,0.898,0.898}}\textbf{75.29} & 49.03 & 42.55 & 44.43   & 41.26 & 59.12          & 58.51          & 54.49          & {\cellcolor[rgb]{0.898,0.898,0.898}}\textbf{61.11}  \\
                                                                        & 10                   & 73.68 & 83.11 & 79.84   & 84.76 & 78.54 & 77.69 & 79.72    & {\cellcolor[rgb]{0.898,0.898,0.898}}\textbf{88.91} & 62.55 & 55.94 & 57.86   & 54.00 & 69.34          & 67.23          & 63.83          & {\cellcolor[rgb]{0.898,0.898,0.898}}\textbf{72.78}  \\ 
\midrule
\multirow{3}{*}{DTD}                                                    & 1                    & 37.94 & 45.93 & 40.19   & 44.39 & 44.78 & 40.94 & 43.91    & {\cellcolor[rgb]{0.898,0.898,0.898}}\textbf{47.10} & 29.49 & 18.32 & 18.55   & 20.39 & 29.69          & 29.17          & 29.20          & {\cellcolor[rgb]{0.898,0.898,0.898}}\textbf{30.73}  \\
                                                                        & 5                    & 72.10 & 76.78 & 71.81   & 76.25 & 74.91 & 73.92 & 70.24    & {\cellcolor[rgb]{0.898,0.898,0.898}}\textbf{78.60} & 58.75 & 41.96 & 41.66   & 46.99 & \textbf{58.82} & 56.76          & 57.19          & {\cellcolor[rgb]{0.898,0.898,0.898}}58.22           \\
                                                                        & 10                   & 82.62 & 87.89 & 83.16   & 86.35 & 85.36 & 83.58 & 84.84    & {\cellcolor[rgb]{0.898,0.898,0.898}}\textbf{89.17} & 71.16 & 54.14 & 53.66   & 60.64 & \textbf{71.36} & 70.41          & 70.33          & {\cellcolor[rgb]{0.898,0.898,0.898}}69.98           \\ 
\midrule
\multirow{3}{*}{UCF101}                                                 & 1                    & 15.49 & 18.53 & 16.26   & 19.60 & 16.89 & 17.82 & 18.87    & {\cellcolor[rgb]{0.898,0.898,0.898}}\textbf{21.12} & 12.34 & 10.65 & 10.49   & 11.62 & 13.01          & 13.73          & 15.49          & {\cellcolor[rgb]{0.898,0.898,0.898}}\textbf{18.62}  \\
                                                                        & 5                    & 47.34 & 54.91 & 49.83   & 56.36 & 49.10 & 50.67 & 52.29    & {\cellcolor[rgb]{0.898,0.898,0.898}}\textbf{57.28} & 38.30 & 34.02 & 34.97   & 36.46 & 38.15          & 40.73          & 45.11          & {\cellcolor[rgb]{0.898,0.898,0.898}}\textbf{48.72}  \\
                                                                        & 10                   & 63.83 & 72.85 & 67.33   & 73.26 & 69.15 & 69.22 & 70.67    & {\cellcolor[rgb]{0.898,0.898,0.898}}\textbf{73.38} & 54.69 & 49.33 & 50.12   & 52.99 & 57.20          & 59.32          & 63.86          & {\cellcolor[rgb]{0.898,0.898,0.898}}\textbf{66.40}  \\ 
\midrule
\multirow{3}{*}{AVG}                                                    & 1                    & 27.04 & 33.75 & 29.20   & 35.42 & 29.98 & 30.62 & 33.68    & {\cellcolor[rgb]{0.898,0.898,0.898}}\textbf{37.65} & 20.62 & 18.26 & 17.34   & 19.12 & 21.59          & 23.02          & 24.63          & {\cellcolor[rgb]{0.898,0.898,0.898}}\textbf{27.25}  \\
                                                                        & 5                    & 54.01 & 63.75 & 57.71   & 65.96 & 57.50 & 59.38 & 62.41    & {\cellcolor[rgb]{0.898,0.898,0.898}}\textbf{68.36} & 45.45 & 41.45 & 40.01   & 43.14 & 46.86          & 48.33          & 51.43          & {\cellcolor[rgb]{0.898,0.898,0.898}}\textbf{54.06}  \\
                                                                        & 10                   & 66.50 & 76.27 & 70.76   & 77.82 & 70.03 & 71.05 & 74.84    & {\cellcolor[rgb]{0.898,0.898,0.898}}\textbf{80.00} & 58.74 & 54.32 & 52.55   & 56.60 & 59.82          & 61.81          & 64.28          & {\cellcolor[rgb]{0.898,0.898,0.898}}\textbf{67.00}  \\
\bottomrule
\end{tabular}}
\label{tab:main_result}
\end{table*}

\section{Fine-class Described Retrieval Dataset}
\label{sec:caption_generation}
Image-Text Retrieval (ITR) in the downstream data domain faces two key challenges: \textbf{(1) Attribute-level precision.} It demands exact alignment of complex visual attributes such as intricate background context. \textbf{(2) Subcategory-level discrimination.} It needs to recognize fine-grained subcategories within a coarse meta-category (e.g., recognizing the Bengal cat or Bombay cat). As shown in Fig.~\ref{fg: dataset_compare}, existing datasets often lack fine-grained subcategory labels; examples of such datasets include MSCOCO~\cite{mscoco} and Flickr30K~\cite{plummer2015flickr30k}. These datasets typically provide general descriptions focusing on common scenes and coarse class names. While some downstream classification datasets are labeled with fine-grained subcategories such as OxfordPets~\cite{pets} and StanfordCars~\cite{cars}. However, they do not contain detailed text descriptions. 
To address these, we construct a benchmark dataset named \textbf{Fine-class Described Retrieval Dataset (FDRD)} based on existing downstream datasets with fine classes.
\textit{To the best of our knowledge, this is the first work that focuses on both fine-grained attribute and subcategory distinctions, aiming to achieve fine-grained image-text matching.}

Specifically, our retrieval dataset consists of image-caption pairs constructed through a semi-automated pipeline, as illustrated in Fig.~\ref{fg: construct_data}. The pipeline comprises:
\begin{enumerate}
    \item \textbf{Caption Draft Generation.} First, we employ InternLM-XComposer~\cite{internlmxcomposer2_5_reward}, an advanced Vision-Language Large Model (VLLM), to automatically generate preliminary captions for images across 9 distinct downstream domains. The specific domain datasets used in this study are introduced in Sec.~\ref{sub: Experimental Settings}. To ensure rich and domain-specific descriptions, we design multiple structured questions tailored to different downstream retrieval scenarios. 

    \item \textbf{Caption Refinement via Feedback Iteration.} Then, the generated drafts are fed into another VLLM (e.g., Qwen2.5-VL~\cite{Qwen2.5-VL}) to evaluate their quality based on the corresponding images. The model provides feedback and updates the draft accordingly, improving it over multiple iterations. 
    
    \item \textbf{Filtering and Post-Processing}. Finally, we manually filter the raw drafts to remove irrelevant information, such as hallucinated content and noise generated by the VLLM. Meanwhile, since the VLLM does not fully recognize fine-grained categories, we manually replace the names of those categories in the captions. Additionally, we standardize the captions and restrict their length to fit within the maximum token length of the CLIP text encoder. 
\end{enumerate}

The final dataset comprises over 1,500 subcategories and 230,000 image-caption pairs, with an average caption length of 47.5 tokens after rare word removal. We plan to publicly release the dataset to facilitate further research in the community. 

\begin{figure*}[t]
\begin{center}
    \includegraphics[width=1\linewidth]{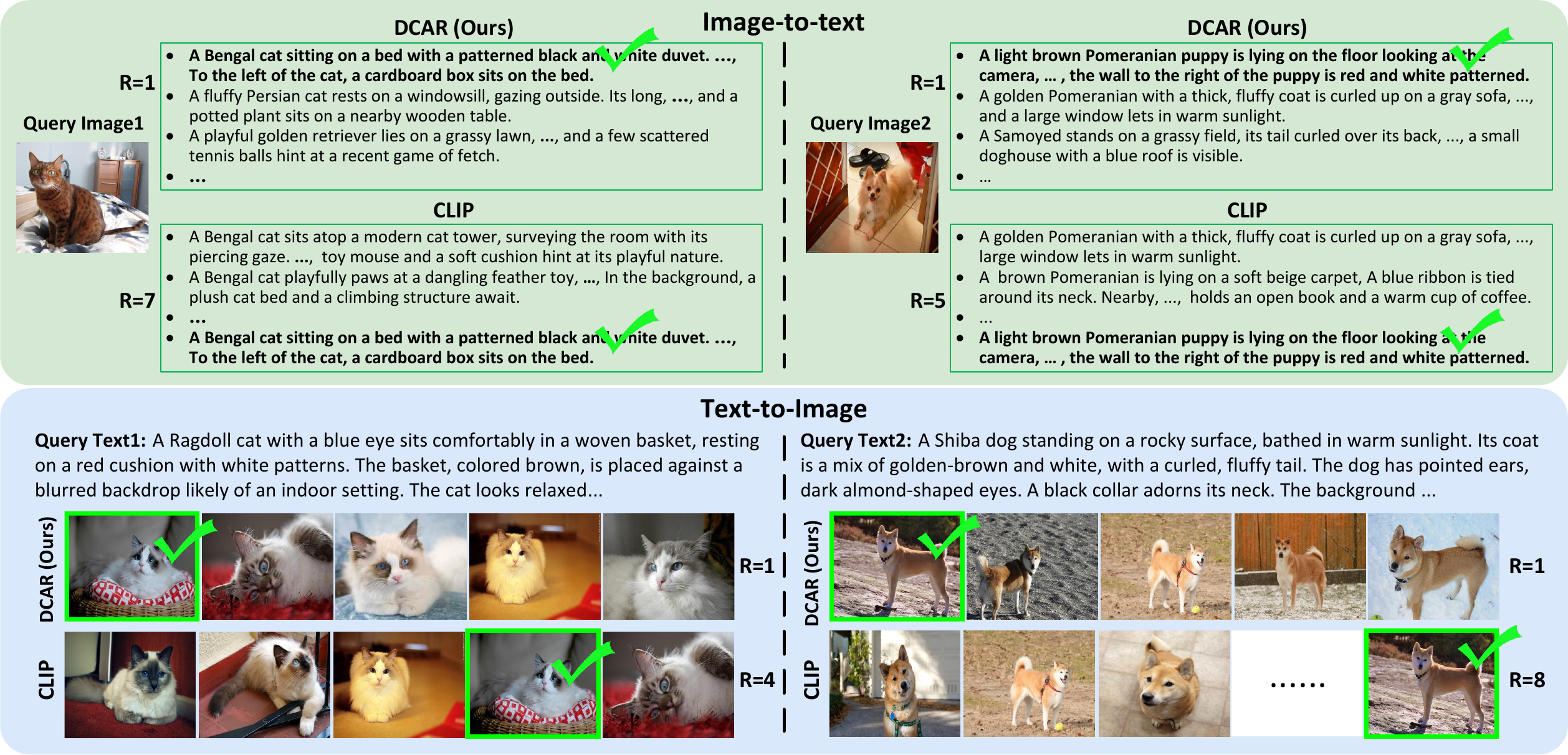}
\end{center}
\caption{Qualitative results of I2T retrieval (top) and T2I retrieval (bottom) by baseline CLIP vs. our method DCAR. The retrieved images (texts) are ranked by retrieval scores, with higher scores appearing at the top. The correct result for every query is marked with a green $\checkmark$, and R denotes the rank at which the correct result is retrieved.}
\label{fg: Qualitative_results}
\end{figure*}

\section{Experiments}
\label{sec:exper}
In this section, we evaluate the effectiveness of our DCAR. We first introduce datasets, baselines, and implementation details. Next, we compare the DCAR with several strong baselines on the FDRD for the Image-Text Retrieval (ITR) task. Finally, we conduct ablation studies to further analyze our method.

\subsection{Experimental Settings}
\label{sub: Experimental Settings}

\quad \textbf{Datasets.}
We evaluate our method on the FDRD constructed in Sec.~\ref{sec:caption_generation}. Specifically, it contains captions for fine-grained images across 9 distinct downstream domains, including Caltech101~\cite{fei2004learning}, Flowers102~\cite{nilsback2008automated}, Food101~\cite{bossard2014food}, OxfordPets~\cite{pets}, StanfordCars~\cite{cars}, DTD~\cite{cimpoi2014describing}, SUN397~\cite{xiao2010sun}, UCF101~\cite{soomro2012ucf101}, and FGVCAircraft~\cite{maji2013fine}. 

\textbf{Baselines.}
To validate the effectiveness of DCAR, we use the following baselines: (1) zero-shot CLIP~\cite{CLIP}, (2) textual prompt learning method: CoOp~\cite{CoOp}, (3) fine-tunes lightweight adapters on top of CLIP: CLIP-Adapter~\cite{Clip-adapter}, (4) multimodal prompt learning approach: MaPle~\cite{MaPLe}, and some advanced image-text retrieval approaches based on CLIP, (5) self-distillation scheme: FILIP~\cite{FILIP}, (6) adaptive contrastive loss optimization: Alip~\cite{yang2023alip}, (7) cross-modal late interaction mechanism: FineCLIP~\cite{FineCLIP}.

\textbf{Evaluation Metrics.} 
For image-to-text, given a query image, the goal is to locate the corresponding caption in the database. Conversely, in text-to-image, a caption is used to find the matching image. We evaluate performance using $Recall@K$ ($R@K$, $K = 1, 5, 10$), where $R@K$ represents the percentage of ground-truth matches found within the Top-$K$ retrieved results. To ensure fair evaluation, each query has exactly one correct match in the database.

\textbf{Experimental details.} We use CLIP-ViT-B/32 as the pre-trained model, freezing the parameters of both the image and text encoders across all methods. The only trainable components are the category awareness network and the vision \& text prompts. All methods are evaluated over three independent runs (using different random seeds), and the average performance is reported. Experiments are conducted under 1, 2, 4, 8, and 16 few-shot settings for all methods, with training, validation, and test image splits kept consistent with the baselines~\cite{CoOp,Clip-adapter,MaPLe}. 
The fine-tuning is performed using image-caption pairs from our FDRD dataset. 
Training is end-to-end on a single NVIDIA RTX 4090 Ti GPU using SGD with an initial learning rate of 0.002, decayed via cosine annealing. 
To mitigate the risk of exploding gradients during early training, we apply a warm-up strategy by fixing the learning rate to $1e\text{-}5$ for the first epoch.

\subsection{Main Results Analysis}
\label{sub: Main Results}
We compared our DCAR with several existing approaches. As shown in Tab.~\ref{tab:main_result}, DCAR achieves the highest average accuracy across all 9 downstream domain retrieval data. 
In the 16-shot setting, DCAR shows strong performance, indicating that the learned embeddings remain meaningful even with limited training samples.

\textbf{Comparison with strong baselines.} On average, DCAR improves over the base CLIP model by +10.61\% and +6.63\% on I2T (Image-to-Text) and T2I (Text-to-Image) retrieval at Recall@1, respectively. Compared to prompt learning methods like CoOp, DCAR achieves +8.99\% on T2I Recall@1, revealing that CoOp may overfit to unimodal prompts, thus limiting CLIP's original cross-modal capabilities. Our dual-prompt framework improves image-text alignment. DCAR outperforms MaPle and CLIP-Adapter across all domains, aided by dynamic token re-weighting, which reduces noise in caption.
For retrieval methods, DCAR surpasses FILIP by +7.67$\%$ (I2T) and +5.66$\%$ (T2I) on average Recall@1. These results indicate that FILIP's token-wise contrastive learning, while effective in general, is not sufficient for ITR tasks where subcategory-level and attribute-level distinctions are critical. 
While FineCLIP performs well, our DCAR still leads by +3.97$\%$ (I2T) and +2.62$\%$ (T2I), proving its superiority in aligning global and local semantics.

\textbf{Cross-domain adaptation.} 
Our DCAR shows robustness across various domains, indicating a well-regularized model that avoids overfitting. 
Notably, our DCAR achieves high recall scores on challenging datasets like Food101 and Flowers102, highlighting its ability to handle fine-grained visual distinctions. 
However, performance is slightly lower on Caltech101 and DTD. Based on our analysis, this may be attributed to the nature of these datasets. DTD is a texture-based dataset, while DCAR is tailored to enhance subcategory-level discrimination, and its effectiveness is limited in such texture recognition scenarios. For Caltech101, although DCAR achieves the best Recall@10 performance, the improvement is marginal due to its broad category diversity, where fine-grained recognition is less crucial and method differences are minimal. 
This suggests our token re-weighting strategy is less effective for coarse-grained tasks. Future work will explore improvements for domains with texture patterns and broad category distributions.

\begin{table}
\centering
\caption {Contribution of each component of our method. The baseline is CLIP. DP: Dual Prompt Learning; CA: Category Aware Augmentation; TW: Dynamic Token Re-Weighting.}
\setlength{\tabcolsep}{0.9mm}{
\begin{tabular}{ccc|c|c|c|c} 
\toprule
                                               &              &              & \multicolumn{2}{c|}{OxfordPets}                   & \multicolumn{2}{c}{Caltech101}                \\ 
\cmidrule{4-7}
DP                                             & CA           & TW           & I2T R@1                 & T2I R@1                 & I2T R@1        & \multicolumn{1}{l}{T2I R@1}  \\ 
\midrule
                                               &              &              & 40.77                   & 30.69                   & 38.99          & 32.17                        \\
$\checkmark$                                   &              &              & 48.88                   & 36.10                   & 44.15          & 33.88                        \\
$\checkmark$                                   & $\checkmark$ &              & 50.94                   & 37.16                   & 46.12          & 34.59                        \\
$\checkmark$                                   &              & $\checkmark$ & 51.04                   & 38.44                   & 47.35          & 36.32                        \\
\rowcolor[rgb]{0.898,0.898,0.898} $\checkmark$ & $\checkmark$ & $\checkmark$ & \textbf{\textbf{52.88}} & \textbf{\textbf{39.82}} & \textbf{48.13} & \textbf{36.78}               \\
\bottomrule
\end{tabular}}
\label{tab:ab_exper}
\end{table}

\begin{figure}[t]
\begin{center}
    \includegraphics[width=1\linewidth]{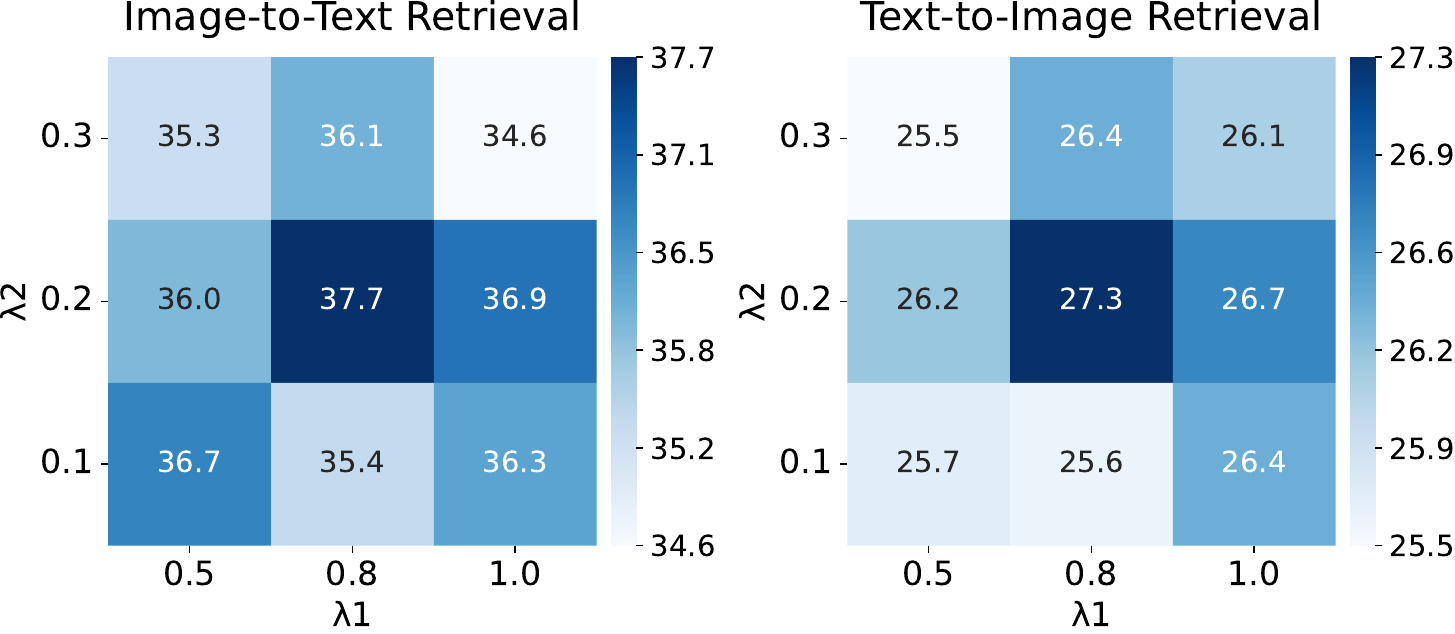}
\end{center}
\caption{Ablation on the parameters $\lambda_1$ and $\lambda_2$. We evaluate the average Image-Text Retrieval Recall@1 across 9 downstream domains in our FDRD dataset.}
\label{fg: para_ana}
\end{figure}

\subsection{Ablation Studies} 
\textbf{Ablation on each component of our method.} As shown in Tab.~\ref{tab:ab_exper}, we conduct ablation studies on Caltech101 (for common-class captions) and OxfordPets (for fine-grained-class captions) image-caption pairs. We observe that each component contributes to retrieval performance, consistently improving over the baseline. The best results are achieved when all components are combined. The use of DP alone boosts I2T performance by +8.11$\%$ and +5.16$\%$ on average across these two datasets, owing to its enhancement of cross-modal alignment. 

\textbf{Ablation on the parameters $\lambda_1$ and $\lambda_2$.}
The parameters $\lambda_1$ and $\lambda_2$ control the weighting of contrastive loss and category-sensitive loss, respectively. In Fig.~\ref{fg: para_ana}, we present the average Recall@1 scores for I2T and T2I retrieval across the 9 downstream domains under different parameter settings. We find that DCAR achieves optimal performance on both I2T and T2I retrieval tasks when $\lambda_1=0.8$ and $\lambda_2=0.2$.

\begin{figure}[t]
\begin{center}
    \includegraphics[width=1\linewidth]{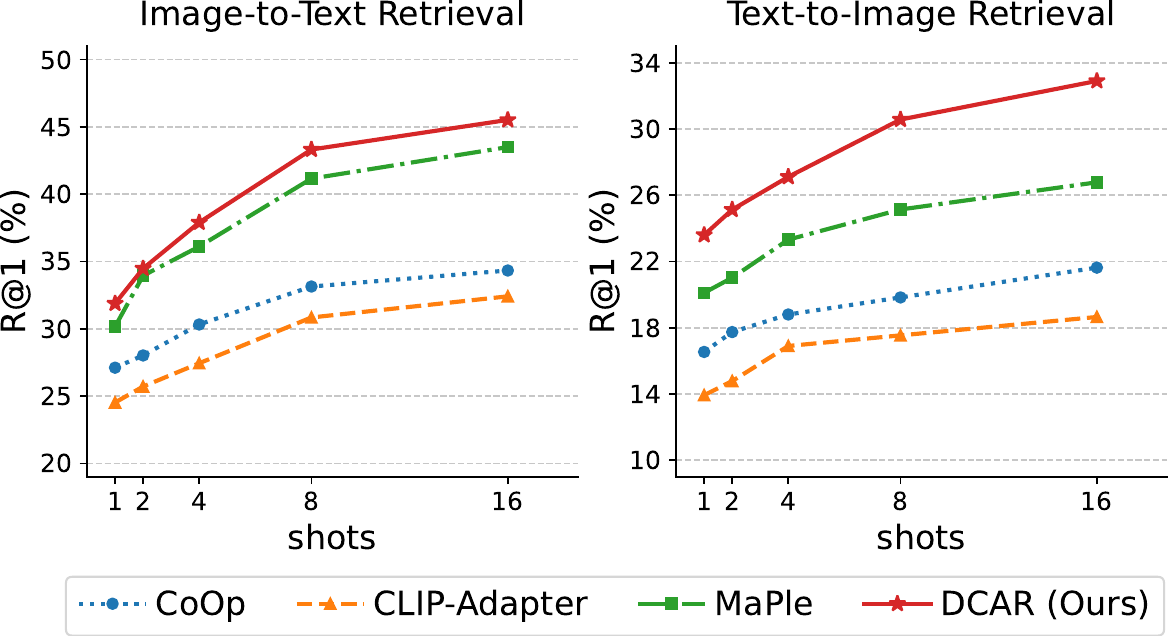}
\end{center}
\caption{Recall@1 results under various few-shot settings for the StanfordCars domain in our FDRD dataset.}
\label{fg: few-shot}
\end{figure}

\subsection{Further Analysis}
\textbf{Qualitative results of Image-Text retrieval.} In Fig.~\ref{fg: Qualitative_results}, we query the same content using both DCAR and the baseline CLIP, followed by a systematic comparison and analysis of their respective retrieval results. For I2T (upper), our method exhibits enhanced reasoning by attending to image-relevant textual tokens (e.g., "a patterned black and white duvet"). For T2I (bottom), DCAR successfully distinguishes between visually similar subcategories. For instance, it correctly retrieves "Ragdoll" vs. "Birman", while CLIP tends to confuse them due to similar fur. Our Top-5 results show stronger semantic relevance to queries, whereas CLIP's outputs exhibit gradual semantic drift. 

\textbf{Various few-shot settings.} In Fig.~\ref{fg: few-shot}, Our method outperforms all other methods on Recall@1 at each shot setting. As the number of shots increases, DCAR continues to deliver significant improvements over the other methods. The category aware augmentation module mitigates coarse-grained feature aggregation in low-shot regimes; the dynamic token re-weighting mechanism prevents overfitting to dominant visual patterns as training samples increase.

\section{Conclusion}
\label{sec:con}
In this work, we identify the limitations of adapting pre-trained vision-language models to the downstream image-text retrieval task and propose a novel dual-prompt framework, DCAR, to address them. By incorporating both attribute-level and subcategory-aware learning, DCAR captures subtle visual semantics and fine-grained distinctions that are often neglected by prior work. We also introduce FDRD, a new benchmark dataset across diverse domains, to evaluate DCAR in challenging retrieval tasks.
Extensive experiments show that DCAR outperforms existing baselines, effectively bridging the modality gap while maintaining fine-grained alignment. DCAR provides a promising direction for extending prompt learning beyond classification, especially in domains that demand nuanced visual reasoning. In future work, we will extend our approach to multi-modal reasoning tasks and further enhance its robustness across diverse domains.

\begin{acks}
This work was supported by the National Major Scientific Instruments and Equipments Development Project of National Natural Science Foundation of China under Grant 62427820, the Fundamental Research Funds for the Central Universities under Grant 1082204112364, the Science Fund for Creative Research Groups of Sichuan Province Natural Science Foundation under Grant 2024NSFTD0035, the Natural Science Foundation of Sichuan Province under grant 2024NSFSC1462 and the Natural Science Foundation of Sichuan under Grant 24NSFSC3404. We would like to thank the anonymous reviewers for their insightful comments and feedback.
\end{acks}

\bibliographystyle{ACM-Reference-Format}
\balance
\bibliography{sample-base}

\end{document}